\documentclass[acmtog, authorversion]{acmart}

\usepackage{subfigure}
\AtBeginDocument{%
  \providecommand\BibTeX{{%
    \normalfont B\kern-0.5em{\scshape i\kern-0.25em b}\kern-0.8em\TeX}}}

\begin{document}

\title{Deep Learning-Based Feature-Aware Data Modeling for Complex Physics Simulations}


\author{Qun Liu}
\email{qliu14@lsu.edu}
\author{Subhashis Hazarika}
\email{hazarika.3@osu.edu}
\author{John M. Patchett}
\email{patchett@lanl.gov}
\author{James Paul Ahrens}
\email{ahrens@lanl.gov}
\author{Ayan Biswas}
\email{ayan@lanl.gov}

\affiliation{%
  \institution{\\Louisiana State University; The Ohio State University; Los Alamos National Laboratory}
}








\renewcommand{\shortauthors}{Qun Liu et al.}

\maketitle

\section{Abstract}


Data modeling and reduction for in situ is important. Feature-driven methods for in situ data analysis and reduction are a priority for future exascale machines as there are currently very few such methods. We investigate a deep-learning based workflow that targets in situ data processing using autoencoders. We propose a Residual Autoencoder integrated Residual in Residual Dense Block (RRDB) \cite{zhang2018residual} to obtain better performance. Our proposed framework compressed our test data into $66$ KB from $2.1$ MB per 3D volume timestep.

\section{Introduction}

Over the last few years, deep learning-based models have become increasingly popular in solving some complex problems from the field of computer vision \cite{akhtar2018threat, liu2018unsupervised}, speech processing~\cite{DeepSpeech} and natural language processing~\cite{DeepNLP}. Recently, there is a growing interest in the scientific data analysis and visualization community to incorporate such powerful machine learning models to solve some of the challenging domain-specific problems. Hong et al.~\cite{lstm_accesspattern} used Long Short-Term Memory (LSTM)~\cite{lstm} based Recurrent Neural Network (RNN) models to estimate the access patterns for parallel particle tracing in distributed computing environments. Han et al.~\cite{flownet} proposed an Autoencoder~\cite{ae} based framework to cluster streamlines and streamsurfaces. Xie et al.~\cite{Xie2018AVA} used neural network embeddings to detect anomalous executions in high performance computing applications. Berger et al.~\cite{GANVolVIS} proposed a Generative Adversarial Networks (GAN)~\cite{GANpaper} based model to synthesize volume rendering images.

We investigate autoencoders (AE) for a particle dataset generated using the Multiphase Flow with Interphase eXchanges (MFiX) simulation code. Our feature preserving data reduction is focused on preserving gas bubbles formed in a fluid, which are the science features of interest. 
We present a Residual Autoencoder, a framework integrated Residual in Residual Dense Block (RRDB) \cite{zhang2018residual} with deep convolutional autoencoder for in situ data reduction. To the best of our knowledge, this is the first work to use RRDB for data reduction purpose and for future in situ deployment.

\begin{figure}[t]
\centering
\includegraphics[width=0.45\textwidth]{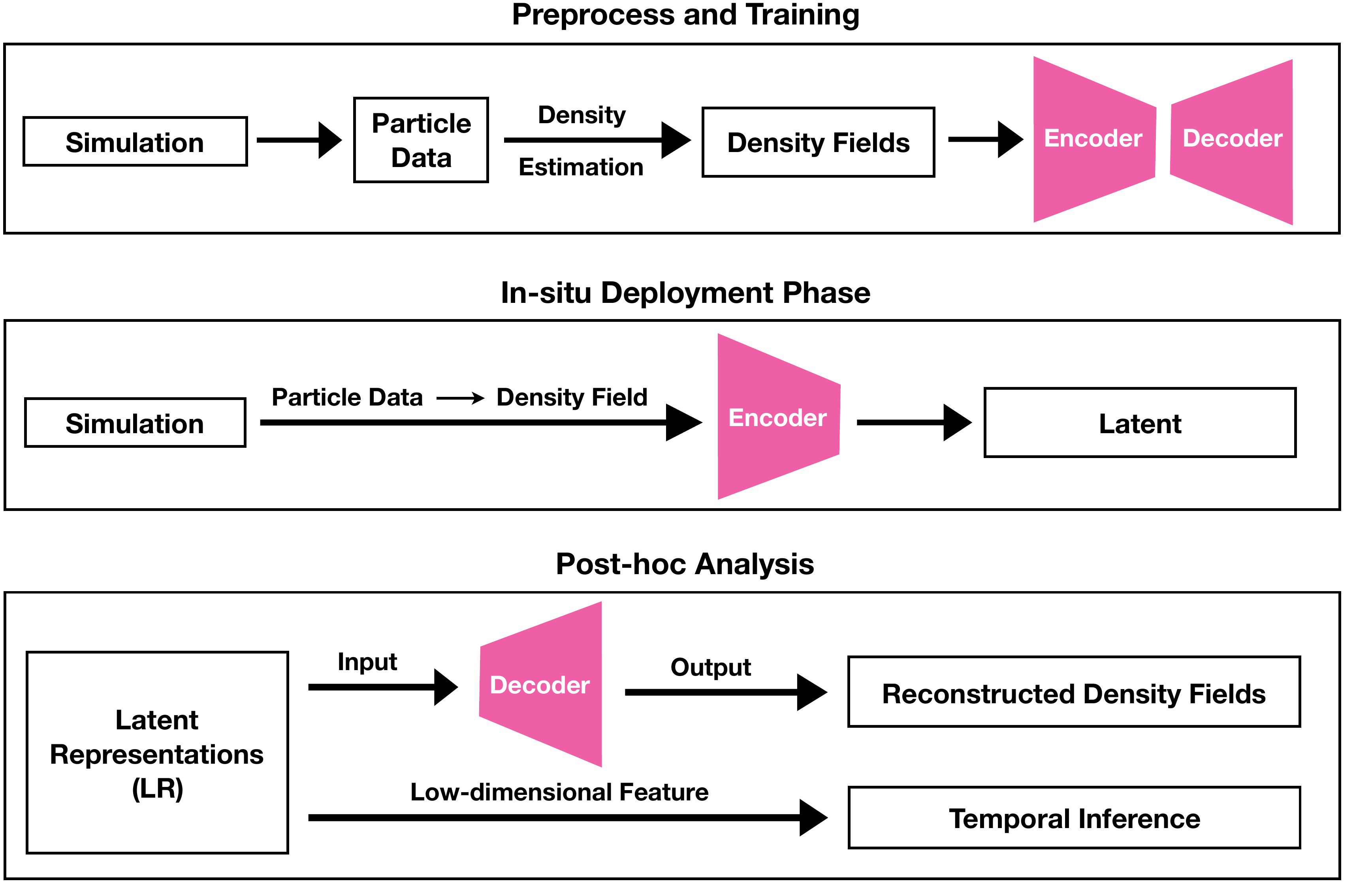}
\caption{The proposed workflow of our framework with in situ.}
\label{fig:wf}
\end{figure}

\section{Architecture}

Our experimental system encodes $128 \times 128 \times 1$ data to latent $16 \times 16 \times 4$, then decodes to produce a $128 \times 128 \times 1$. 
The designed architecture of 
Residual Autoencoder
in situ have shown in Fig. \ref{fig:bf}. 
The kernel size for all convolutional layers we set to $3 \times 3$, and $2 \times 2$ as the size for all MaxPooling and UpSampling layers. All convolutional layers have $64$ filters except the last two RRDBs in the encoder which its last convolution layers have $32$ and $4$ filters.

\section{Methodology}

\begin{figure}[b]
\centering
\includegraphics[width=0.5\textwidth]{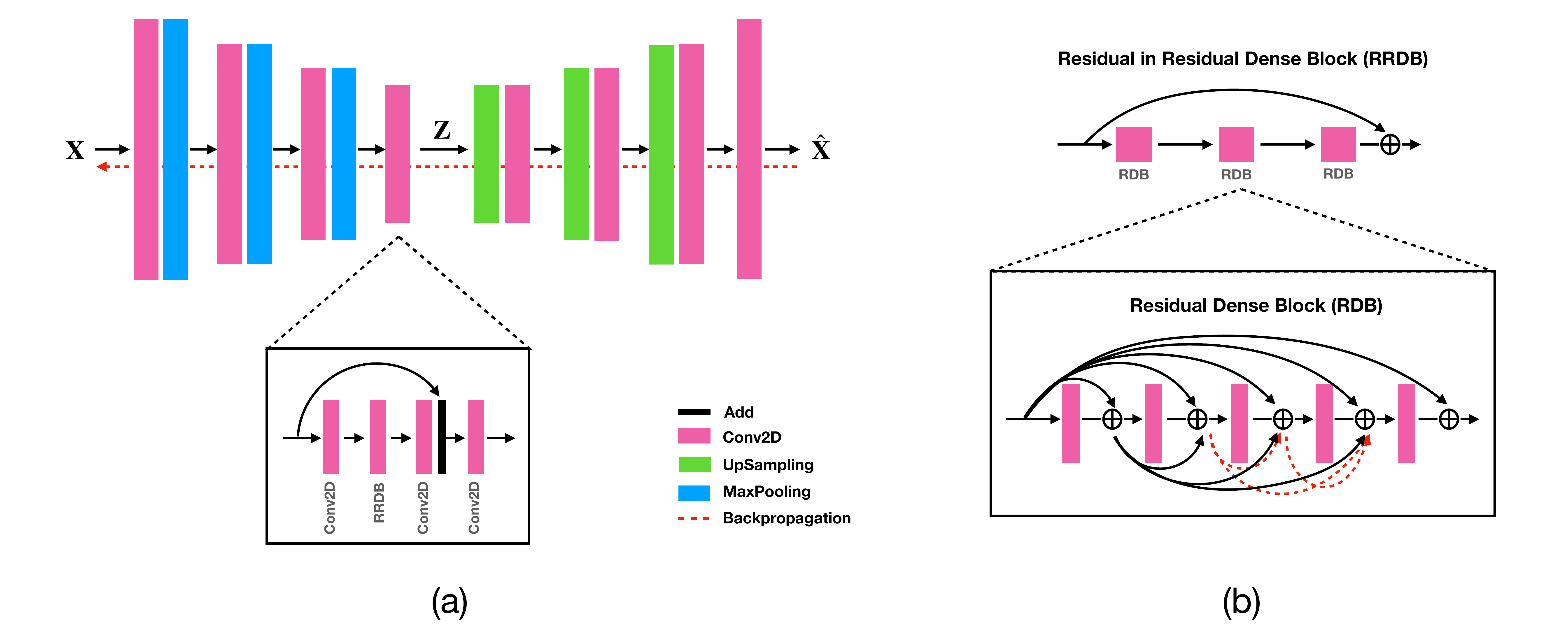}
\caption{Overview architecture of our proposed framework.}
\label{fig:bf}
\end{figure}

Autoencoders are known for dimensionality reduction and unsupervised feature learning. They consist of two parts. The encoder uses a simple neural network, produces a representation $Z$ of the input data $X$. The decoder, also using a simple neural network, reverses the encoding, converting $Z$ to $\hat{X}$ as seen in Fig. \ref{fig:bf}.  
Our work uses a convolutional autoencoder or CAE. A CAE replaces a simple neural network with multiple convolutional and deconvolutional layers for both the encoder and decoder, respectively.
The Residual Dense Block (RDB) \cite{zhang2018residual} we used in our architecture has shown in the bottom in Fig. \ref{fig:bf}(b). RDB facilitates the enriched local and hierarchical features learning through dense convolutional layers resulting in a contiguous memory (CM) mechanism while stabilizes the network training.

\section{Evaluation}
 
\begin{figure}[t]
\centering
\subfigure[Normal CAE]{
    \includegraphics[width=0.45\linewidth, keepaspectratio]{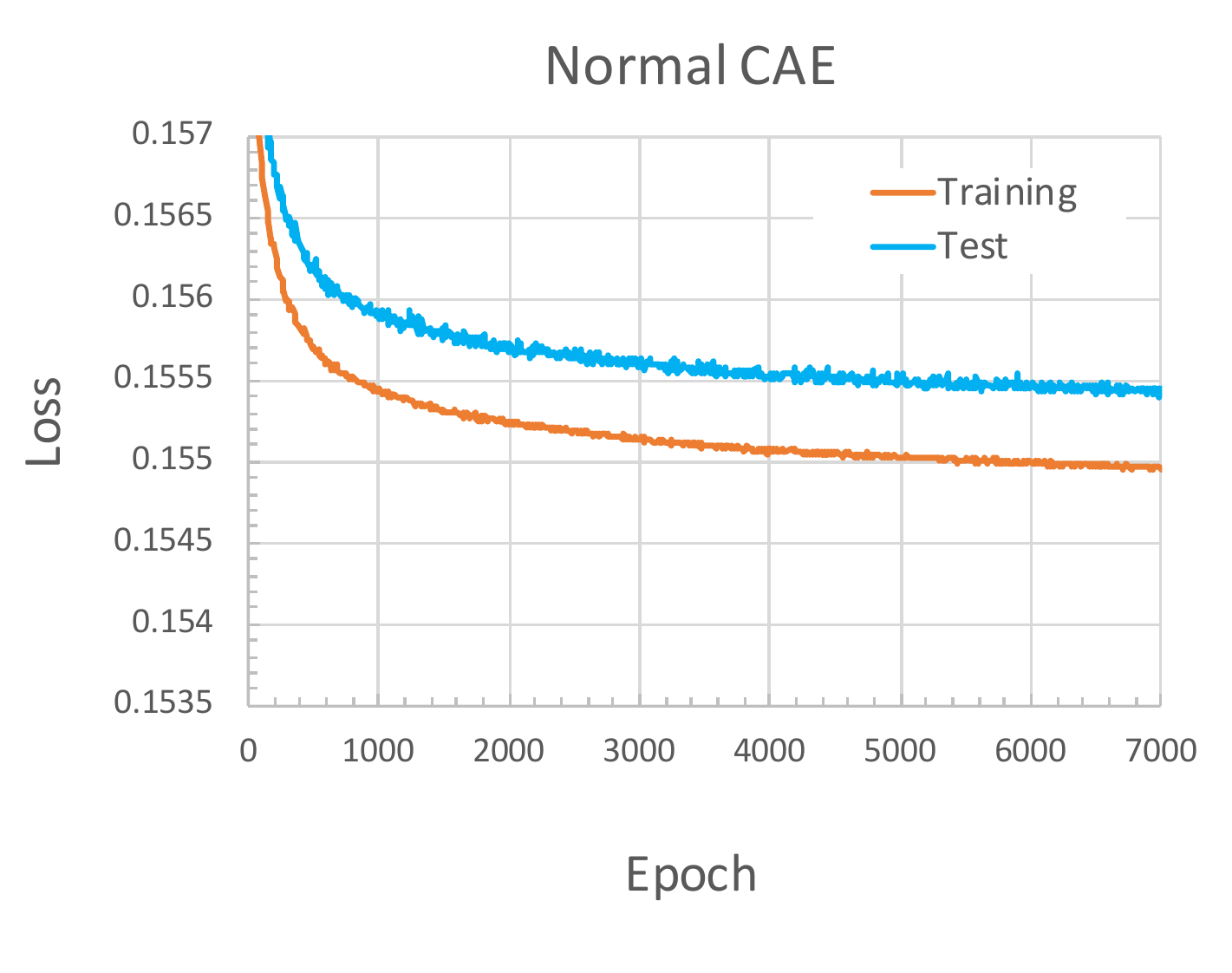}
    \label{fig:cae_loss}
}
\subfigure[Our framework]{
    \includegraphics[width=0.45\linewidth, keepaspectratio]{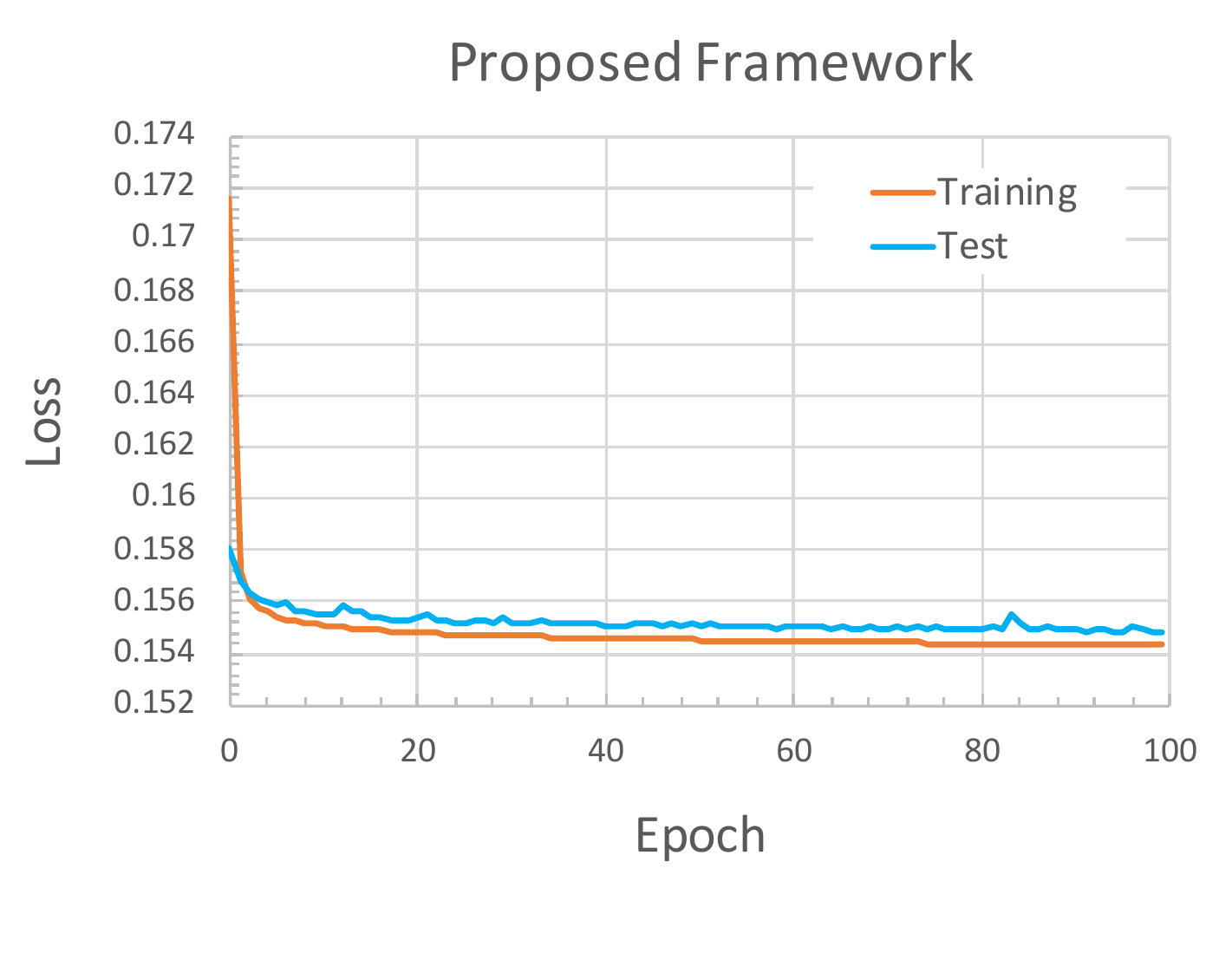}
    \label{fig:ra_loss}
}


\caption[Optional caption for list of figures]{Losses in training and testing.}
\label{fig:loss}
\end{figure}


\begin{figure}[b]
\centering
\includegraphics[width=0.45\textwidth]{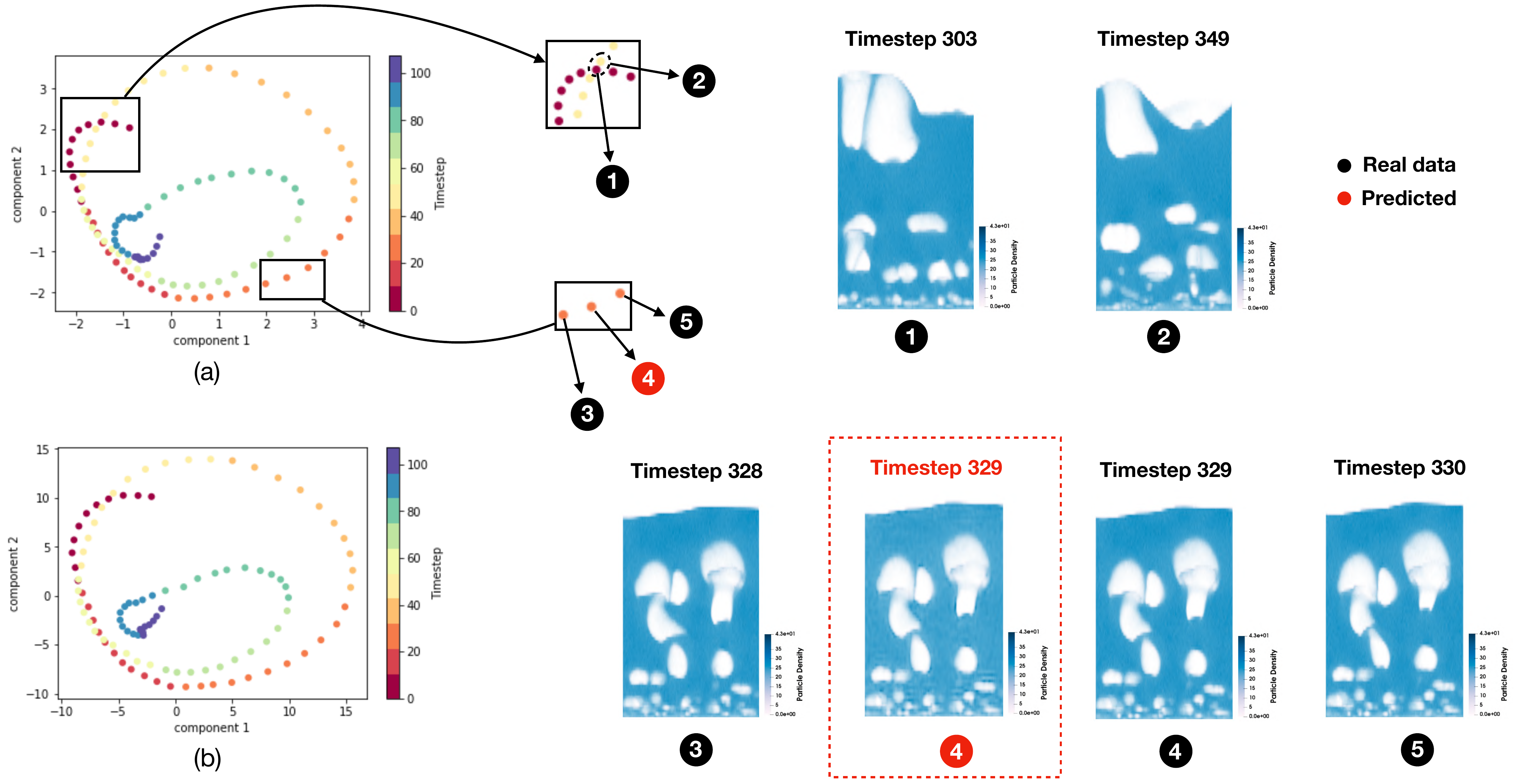}
\caption{Visualization and analysis on principal component analysis (PCA) results.}
\label{fig:pca_results}
\end{figure}

\textbf{Dataset} Our experimental data was generated from an MFiX carbon particle simulation. We initially preprocessed the raw particle data into a density field for use as our experimental dataset to explore our proposed framework. It is comprised of 409 timesteps, each timestep is a 3D volume, $128 \times 16 \times 128$. We removed the first 60 timesteps from the dataset as they were part of the initialization. We then selected 241 timesteps as training data and used the remaining 108 timesteps as test data.

 The training and test process regarding the losses are shown in Fig. \ref{fig:loss}.
We can see from Fig. \ref{fig:cae_loss} that the training loss for the normal CAE reached around $0.1550$ with training $7000$ epochs and the test loss is around $0.1555$. From Fig. \ref{fig:ra_loss} we can see the training loss for our proposed framework achieved around $0.1543$ within $100$ epochs and the test loss is around $0.1548$. Both Fig. \ref{fig:cae_loss} and Fig. \ref{fig:ra_loss} showed the best epochs which thereafter the losses will not decrease any more. Thus, we adopted early stopping strategy in case of overfitting problem. The training for our proposed framework is more stable than the normal CAE and obtained fast converge and best performance with much less training epochs.

\begin{figure}
\centering
\includegraphics[width=0.45\textwidth]{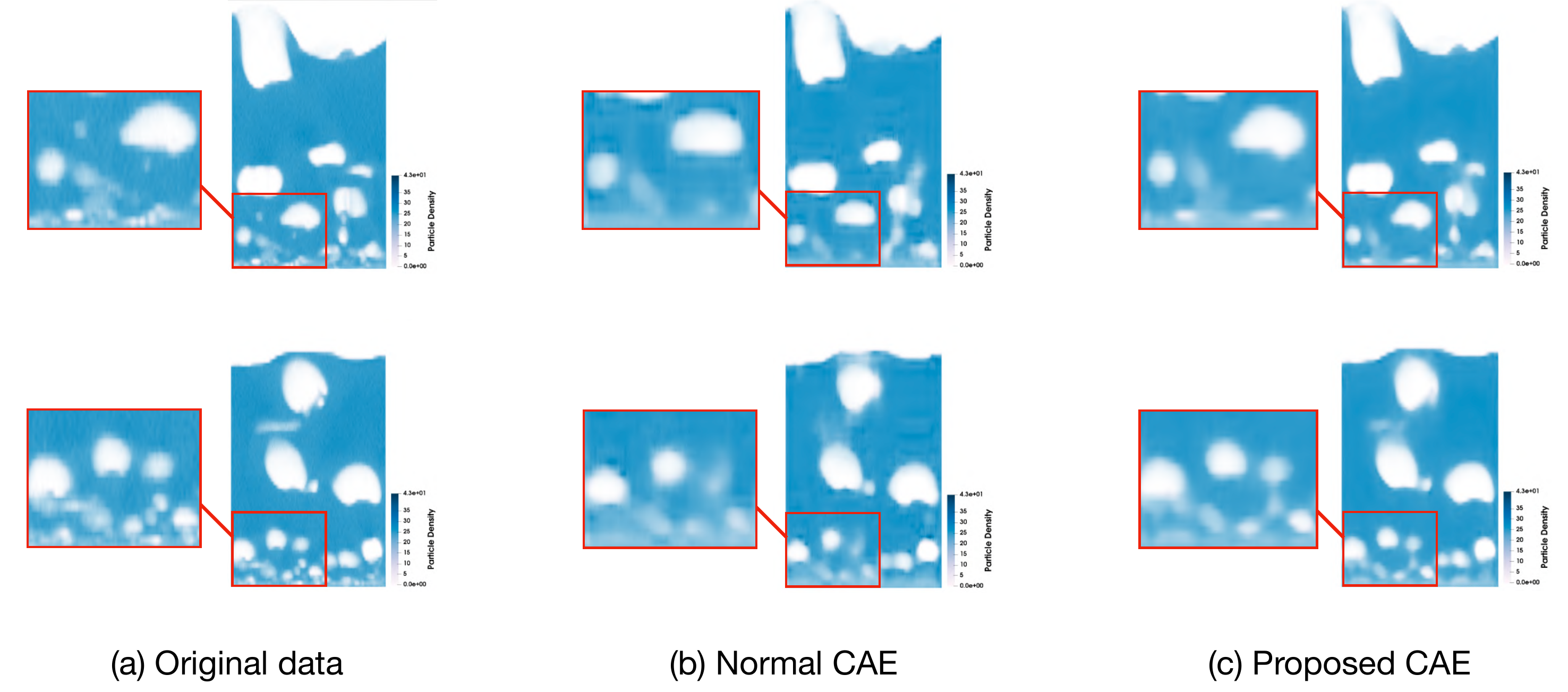}
\caption{Visualization of output. From top to bottom: timesteps 351 and 401.}
\label{fig:all_in_one}
\end{figure}

The proposed framework has to be pretrained before being deployed in situ. The offline training information has been provided in Table \ref{table:offline}. Since our proposed framework consists of multiple RRDBs in the encoder and decoder, it resulted in a very large network having a large amount of parameters needing much more time to train, but it achieved better performance as shown in Fig. \ref{fig:all_in_one} and Table \ref{table:offline}. We trained and tested our framework using $4$ GeForce GTX 1080 Ti with compute capability 6.1, the average processing time on test data for encoding and decoding are shown in Table \ref{table:process}.

We have visualized the test data and its latent representations in 2D using Principal Component Analysis (PCA) as shown in Fig. \ref{fig:pca_results}(a) and Fig. \ref{fig:pca_results}(b). We can see that the learned latent representations have very similar distribution with test data and in the future the latent space learned from our framework enables the possibilities for feature analysis and feature tracking, further research could be investigated such as latent space interpolation for time-step selection as illustrated in Fig. \ref{fig:pca_results}, since the linear interpolation can be performed in the PCA space (red dot interpolated using its two neighbour black dots) and using inverse PCA, we can return to original encoded space before using the decoder.


\begin{table}[h]
\begin{center}
\begin{tabular}{p{1cm}|p{1.1cm}|p{1.1cm}|p{1.1cm}|p{1.1cm}|p{1cm}}
\hline
 \textbf{Net} & \textbf{Training Time(h)} & \textbf{Training Loss} & \textbf{Testing \newline Loss} & \textbf{Total \newline Params} & \textbf{Weight Size}     \\ \hline
    CAE         & 1.8128   &  0.1551  &  0.1555  &  4385  & 50KB \\
    Ours & 32.2122   &  0.1543  &  0.1549  &  24,638,885   & 99.1MB \\
\hline 
\end{tabular}
\end{center}
\caption{Offline training information}
\label{table:offline}
\end{table}

\vspace{-25pt}

\begin{table}[h]
\begin{center}
\begin{tabular}{p{2cm}|p{2cm}|p{1.5cm}|p{1.5cm}}
\hline
 \textbf{Encoding} & \textbf{Decoding} & \textbf{Original I/O} & \textbf{Encoding w. I/O} \\ \hline
    0.3074 &  0.1273  & 0.1849   &  0.3818   \\
\hline 
\end{tabular}
\end{center}
\caption{Average processing time (in seconds) of proposed framework for one timestep}
\label{table:process}
\end{table}

\vspace{-20pt}

\section{Conclusion}

We presented a potential workflow for in situ data modeling based on deep learning, and further we made initial efforts on feature-driven data reduction for in situ by integrating RRDB in CAE to improve performance and thus proposed a Residual Autoencoder. Our proposed framework can effectively and efficiently compress data as demonstrated in our experiments.


\bibliographystyle{ACM-Reference-Format}
\bibliography{main,main1}

\end{document}